\documentclass[twoside]{article}

\usepackage[accepted]{paperstyle}
\usepackage{amssymb}
\usepackage{graphicx} 
\usepackage{multirow} 
\usepackage{multicol} 
\usepackage{booktabs}
\usepackage{natbib}
\setcitestyle{authoryear,round}

\begin{document}

%

%

\twocolumn[

\aistatstitle{Type-based Neural Link Prediction Adapter for Complex Query Answering}

\aistatsauthor{ Lingning Song \And Yi Zu \And  Shan Lu \And Jieyue He}

\aistatsaddress{ Southeast University \And Southeast University \And Nanjing Fenghuo Tiandi \\
Communication Technology \\
Co., Ltd \And Southeast University} ]

\begin{abstract}
Answering complex logical queries on incomplete knowledge graphs (KGs) is a fundamental and challenging task in multi-hop reasoning. Recent work defines this task as an end-to-end optimization problem, which significantly reduces the training cost and enhances the generalization of the model by a pretrained link predictors for query answering. However, most existing proposals ignore the critical semantic knowledge inherently available in KGs, such as type information, which could help answer complex logical queries. To this end, we propose TypE-based Neural Link Prediction Adapter (TENLPA), a novel model that constructs type-based entity-relation graphs to discover the latent relationships between entities and relations by leveraging type information in KGs. Meanwhile, in order to effectively combine type information with complex logical queries, an adaptive learning mechanism is introduced, which is trained by back-propagating during the complex query answering process to achieve adaptive adjustment of neural link predictors. Experiments on 3 standard datasets show that TENLPA model achieves state-of-the-art performance on complex query answering with good generalization and robustness.
\end{abstract}

\section{INTRODUCTION}

Knowledge Graph (KG) is usually a heterogeneous graph that stores structural information and knowledge. The facts in KG are usually represented in the form of triples, which capture many kinds of relationships between entities. In recent years, KGs have been widely used in variety domains, such as question answering, search engines and recommender systems \citep{domain:1,domain:2}. Reasoning on knowledge graphs aims to infer new knowledge or answer queries based on existing knowledge or facts. One important task on KG reasoning is complex logical query answering.

Complex logical queries can be represented with First-Order Logic (FOL), which includes logical operations like existential quantifier ($\exists$), conjunction ($\wedge$), disjunction ($\vee$), and negation ($\neg$). For example, the query "List the research fields of non-Canadian scholars who have won the Turing Award" can be represented as an FOL query:
\begin{equation}
\begin{split}
q=&V_?.\exists V:Win\left(V,Turing\ Award\right) \\
&\wedge\neg Citizen\left(V,Canada\right) \wedge Field \left(V, V_?\right)
\end{split}
\end{equation}
Traditionally, symbolic methods such as fuzzy logic \citep{fuzzy:logic}, logic programming \citep{logic:progarmming} and probabilistic reasoning \citep{probabilistic:reasoning} are utilized to handle the problem of reasoning. These methods traverse the whole graph and extract all possible assignments for intermediate variables, resulting in exceptional interpretability and readability. Symbolic methods are also capable of producing the correct answer when all facts are given on knowledge graphs. However, finite and discrete symbolic representations are not sufficient to describe all intrinsic relationships between data, and struggle to handle duality and noise. Besides, many real-world knowledge graphs are incomplete, which severely limits the usage of symbolic methods on knowledge graphs.

Motivated by the great success of knowledge graph embedding (KGE) on answering one-hop KG queries, a series of work emerges to answer complex logical queries on incomplete knowledge graphs by learning embedding representations for each FOL query \citep{gqe,betae,q2b,cone,fuzzqe,gnnqe,Query2GMM}. These methods embed both entities and FOL queries into the same low-dimensional space and translate logical operations into neural logical operations in the embedding space, facilitating the identification of implicit correlations within the data. Nevertheless, these methods need to be trained on millions of generated complex logical queries, leading to large training time overhead and poor generalization to out-of-distribution (OOD) query structures. Furthermore, they have difficulty in interpreting what the intermediate variables stand for and fail to provide explicit reasoning evidence to explain the results, which is unfavorable for further visual analysis of the reasoning.

To these ends, CQD \citep{cqd} proposes an end-to-end optimization framework, which calculates the truth value of each one-hop atom provided by a pretrained neural link predictor, and finds a set of entity assignments that maximizes the truth score of a FOL query via two optimization solutions. Consequently, it is able to explicitly interpret intermediate variables in queries, and does not require training on complex queries. However, its accuracy is limited due to the approximation during optimization, as the search space is exponential to the number of intermediate variables. QTO \citep{qto} leverages the independence encoded in the tree-like computation Directed Acyclic Graphs (DAGs) of the query to greatly reduce the search space, and thus can efficiently find the theoretically optimal solution. However, these methods ignore critical semantic knowledge inherently available in KGs, such as type information \citep{autoeter,cet}. The integration of type information can enhance the capability of neural link predictors, enabling them to adapt better to complex query answering tasks.

In this paper, we propose \textbf{T}yp\textbf{E}-based \textbf{N}eural \textbf{L}ink \textbf{P}rediction \textbf{A}dapter (TENLPA), which integrates type information in KGs and enhances the prediction performance of the neural link predictor by learning additional adaptive functions, thereby improving its applicability to complex logical query tasks. We first build two type-based entity-relation graphs, which aim to enhance possible associations between entities and relations by type information. In a subsequent step, we perform adaptive calibration on the neural adjacency matrix, in order to convert KGE scores into probabilistic representations via the learnable calibration function. Finally, we use type-based neural link prediction adapter to carry out further adaptive adjustment leveraging type information. Our main contributions can be summarized as follows:
\begin{itemize}
\item We propose a novel type-based neural link prediction adapter that learns the interconnections among entities, relations and types through two adaptive functions, thereby enabling adaptive adjustment of the results obtained by the neural link predictor for answering complex query on KGs.
\item We construct two innovative type-based entity-relation graphs that provide a fresh interpretation of the association between entities and relations from the perspective of types.
\item We evaluate the effectiveness of TENLPA on complex logical query answering on 3 standard datasets. Results demonstrate that our model outperforms existing state-of-the-art methods in answering FOL queries.
\end{itemize}

\section{RELATED WORK}

Our work is related to knowledge graph completion, complex logical query and type-aware embedding models, so we will summarize the existing knowledge graph completion methods, briefly introduce the relevant research progress in complex logical query and discuss type-aware embedding models in this section.

\subsection{Knowledge Graph Completion}
The goal of knowledge graph completion is to infer missing relational links (one-hop queries) on an incomplete KG. To address this issue, a popular approach is knowledge graph embedding \citep{transe:fb15k,distmult,complex,conve,rotate,cone:kge} that learns low-dimensional vectors for each entity and relation, and measures the likelihood of a triplet by a defined scoring function over the corresponding vectors. Rule learning methods \citep{neurallp,drum,rnnlogic} first extract interpretable logic rules from the knowledge graph, which are subsequently employed to predict the links. In addition, some studies adopt graph neural networks \citep{compgcn,nbfnet} to learn the entity or pairwise representations for knowledge graph completion. In this work, a pretrained knowledge graph embedding (KGE) model is used to calculate the truth values of one-hop queries.

\subsection{Complex Logical Query}
Complex logical queries are one-hop KG queries combined by logical operations, which extend knowledge graph completion to predict answer entities for queries with conjunction, disjunction or negation operators. Embedding-based methods represent sets of entities as geometric shapes or probabilistic distributions, and find the solution by obtaining the nearest neighbor entities to the answer set representation. GQE \citep{gqe} can answer conjunctive query ($\wedge$) by representing queries as vectors in the embedding space, which is later extended by Query2Box \citep{q2b} to existential positive first-order (EPFO) queries ($\exists,\wedge,\vee$) and BetaE \citep{betae} to the FOL query ($\exists,\wedge,\vee,\neg$). HypE \citep{hype} and ConE \citep{cone} respectively utilize hyperboloid embeddings and cone embeddings, allowing the operators to attain the desired properties. However, embedding-based methods usually lack interpretability and the quality of set representation may be compromised when the set is large. To address this, some approaches integrate more interpretable fuzzy logic to handle FOL queries. FuzzQE \citep{fuzzqe} improves embedding-based methods with t-norm fuzzy logic, which satisfies the axiomatic system of classical logic. GNN-QE \citep{gnnqe} decomposes the query into relational projections and logical operations over fuzzy sets, and employs a GNN to execute relational projections. Nevertheless, all the above methods require training on complex queries to achieve strong results, which restricts their capacity to generalize to more complicated query structures and prevents their improvement by more powerful knowledge graph completion models. CQD \citep{cqd} defines a complex logical query as an end-to-end optimization problem and uses a pretrained KGE model to infer answers. Two strategies are proposed to approximate the optimal solution, namely CQD-CO that directly optimizes on the continuous embeddings and CQD-Beam that uses beam search to generate a sequence of entity assignments. But the method still suffers from deficiencies in terms of efficiency and accuracy. QTO \citep{qto} can efficiently find the theoretically optimal solution by a forward-backward propagation on the tree-like computation graph, and reduce the search space by utilizing the independence encoded and the divide-and-conquer strategy. However, QTO also has some limitations, such as the fixed KGE calibration function and the ignorance of the importance of type information during reasoning, which lead to the gap between prediction results and answers.

Therefore, we propose TENLPA model, which integrates type information, improves the KGE calibration function and designs a neural link prediction adapter. Through gradient-based optimization trained on complex logical queries, our method can obtain superior performance in answering FOL queries.

\subsection{Type-aware Embedding Models}
Most KGC methods only study the triples independently and ignore the potential and valuable information, which degrades their link prediction performance. To alleviate this issue, several endeavors have been made to optimize the embedding performance by adding auxiliary information, among which type information stands out as the most common choice characterized by its less noise. The supervised approaches \citep{tkrl, tcrl, transt} attempt to learn type representations from annotated type information, requiring explicit entity type. Unsupervised learning methods \citep{autoeter, utci, cet} take one or more type-agnostic KGC models as their base framework and modify the score function to consider entity type compatibility for evaluating fact plausibility. However, these works cannot be directly used for answering FOL queries because of multi-hop reasoning, producing intermediate uncertain entities. TEMP \citep{temp} uses type embeddings to enrich entity and relation representations and can be easily incorporated into existing QE-based models \citep{gqe, q2b, betae} for multi-hop reasoning. Unfortunately, it is not suitable for fuzzy-logic models, which also requires the advantage of type information. Our work captures the connections between entities and relations, which can assist fuzzy-logic models better utilize type information for complex logical queries.

\section{PRELIMINARIES}
Several relevant definitions are given to facilitate the interpretation of the TENLPA model.

\subsection{Knowledge Graphs and Knowledge Graph Embeddings}
Given a set of entities $\mathcal{V}$ and a set of relations $\mathcal{R}$, a knowledge graph $\mathcal{G}$ is defined as $\mathcal{G}=(\mathcal{V},\mathcal{R},\mathcal{T}$) where $\mathcal{T}$ is the set of triplets. Each triplet in $\mathcal{T}$ is a fact $(h,r,t)$, where $h,t\in\mathcal{V}$ are respectively the head and tail entities of the triplet, and $r\in\mathcal{R}$ is the edge of the triplet connecting head and tail. KGE models learn mappings $\mathcal{V}\rightarrow\mathbb{R}^{d_\mathcal{V}},\ \mathcal{R}\rightarrow\mathbb{R}^{d_\mathcal{R}}$ from entities and relations to vectors, and score the likelihood of a triplet $(h,r,t)$ by a function $f_r(h,t)$.

\subsection{First-Order Logic}
Following the BetaE citep{betae}, a FOL query $q$ can be written in its disjunctive norm form (DNF), with which the query can be represented as a disjunction of several conjunctions:
\begin{equation}
q\left[V_?\right]=V_?.\exists V_1,V_2,\ldots,V_k\in\mathcal{V}:c_1\vee c_2\vee\cdots\vee c_n
\end{equation}
where $V_1,V_2,\ldots,V_k$ are variables, $V_?\in\left\{V_1,V_2,\ldots,V_k\right\}$ is the answer variable and $c_n=e_1^n\land\cdots\land e_{m_n}^n$ is a conjunction of several literals. Each literal $e_j^i$ represents the truth value of an atomic relational formula on relation $r$ between two entities or its negation, respectively:
\begin{equation}
e_j^i=
\left\{
\begin{array}{lr}
r(c,V)\ or\ r(V^\prime,V) \\
\neg r(c,V)\ or\ \neg r(V^\prime,V)
\end{array}
\right.
\end{equation}
where $V,V^\prime$ are variables, and $c$ is a constant (anchor) entity.

The goal of query answering is to find a feasible variable assignment that renders $q$ true. We use $e_j^i$ to denote the likelihood that the relationship holds true, whose generalized truth value is in $[0,1]$. Following the QTO \citep{qto}, we formalize this as an optimization problem:
\begin{equation}
\begin{split}
q\left[V_?\right]&=V_?.V_1,\ldots,V_k \\
&=\mathop{\arg{max}}\limits_{V_1,\ldots,V_k\in\mathcal{V}}\left(e_1^1\top\ldots\top e_{m_1}^1\right)\bot\ldots\bot\left(e_1^n\top\ldots\top e_{m_n}^n\right)   
\end{split}
\end{equation}
where $e_j^i\in[0,1]$ is scored by a pretrained KGE model based on the likelihood of the atomic formula.$\top$ and $\bot$ are generalizations of conjunction and disjunction over fuzzy logic on $\left[0,1\right]$, namely \textit{t-norm} and \textit{t-conorm}. In this paper, we use \textit{product t-norm} with its corresponding conorm to implement conjunction and disjunction.

\section{METHODOLOGY}

\begin{figure*}[htb]
    \centering 
    \includegraphics[width=0.9 \linewidth]{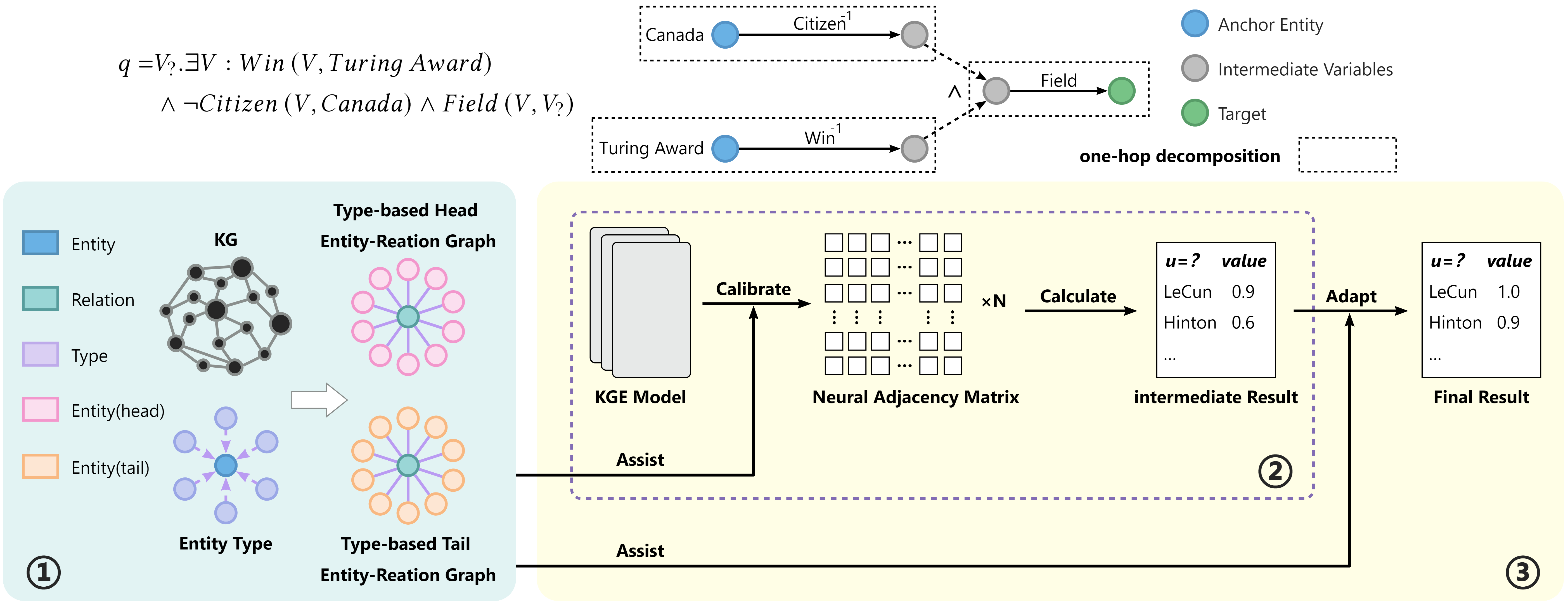}
    \caption{Overview of TENLPA, which consists of three parts.}
    \label{figure1}
\end{figure*}

We propose Type-based Neural Link Prediction Adapter (TENLPA) for complex logical query answering on KGs. TENLPA model is composed of three parts: 1) Two type-based Entity-Relation Graphs, which leverage type information of entities to enrich the potential connections between entities and relations. 2) Adaptive Calibration of Neural Adjacency Matrix, which converts KGE scores into probabilistic representations by a learnable calibration function. 3) Type-based Neural Link Prediction Adapter, which adaptively adjusts prediction results utilizing type information to more effectively complete complex query answering tasks. The overview of TENLPA is shown in Figure \ref{figure1}.

\subsection{Type-based Entity-Relation Graphs Construction}
\label{type:graph}
We formally establish two entity-relation graphs $\mathcal{G}_{hr}$ and $\mathcal{G}_{tr}$ using type information and the original KG. Let $\mathcal{G}=\left(\mathcal{V},\mathcal{R},\mathcal{T}\right)$ be an original KG, $\mathcal{C}=\left\{\left(e,type\right)\mid e\in\mathcal{V},type\in\mathcal{P}\right\}$ be type information. For a relation $r\in\mathcal{R},\mathcal{T}_r\subseteq\mathcal{T}$ denotes the set of triplets in which $r$ occurs. For a triplet $t\in\bigcup_{r\in\mathcal{R}}\mathcal{T}_r$, we denote $t_{hd}$ and $t_{tl}$ as the head and tail entities of $t$, respectively. Moreover, $tp_t\left(t_{hd}\right)=\left\{type\mid\left(t_{hd},type\right)\in\mathcal{C}\right\}$ denotes the set of types of the head of $t$, and $tp_t\left(t_{tl}\right)=\left\{type\mid\left(t_{tl},type \right)\in\mathcal{C}\right\}$ is defined analogously. Since $r$ may occur in multiple triplets, we respectively take the union/intersection of the types of the head/tail entities of triplets in which $r$ occurs. For $r\in\mathcal{R}$, we define the set of types of head/tail entities related to $r$ as:
\begin{equation}
tp_r^{hd}=\bigcup_{t\in\mathcal{T}_r}tp_t\left(t_{hd}\right),\qquad tp_r^{tl}=\bigcap_{t\in\mathcal{T}_r}tp_t\left(t_{tl}\right)
\end{equation}
Let $e_{type}$ represent the type information of $e$, $\mathcal{T}_{hr}=\left\{(e,r)\mid e\in\mathcal{V},e_{type}\in{tp}_r^{hd},r\in\mathcal{R}\right\}$ denotes type-based head entity-relation binary association. We define $\mathcal{G}_{hr}=\left(\mathcal{V},\mathcal{R},\mathcal{T}_{hr}\right)$ as a type-based head entity-relation graph. Similarly, $\mathcal{G}_{tr}=\left(\mathcal{V},\mathcal{R},\mathcal{T}_{tr}\right),\mathcal{T}_{tr}=\left\{(e,r)\mid e\in\mathcal{V},e_{type}\in{tp}_r^{tl},r\in\mathcal{R}\right\}$ is defined as a type-based tail entity-relation graph. Two entity-relation graphs establish new associations between entities and relations through type information, which capture a broader range of potential connections between entities and relations.
\subsection{Neural Adjacency Matrix Calibration}
We define a neural adjacency matrix $\mathbf{M}_r\in[0,1]^{\left| V\right|\times\left| V\right|}$ for each relation $r\in\mathcal{R}$, where $\left(\mathbf{M}_r\right)_{i,j}=p_r\left(e_i,e_j\right)$ is the probability of the triplet $\left(e_i,r,e_j\right)$ being true, and $e_i,e_j\in\mathcal{V}$ are the $i,j$-th entities. We score the likelihood of the triplet $\left(e_i,r,e_j\right)$ via $f_r\left(e_i,e_j\right)$ of the KGE model, and subsequently normalize the score by a calibration function to obtain the corresponding value of each element in $\mathbf{M}_r$. Specifically, since the score provided by KGE model is not a probability, it is imperative to calibrate it to a probability between $[0,1]$. We adopt a monotonically increasing function for calibration, which guarantees that the score is faithfully calibrated to 1 if there is an edge $r$ between $e_i$ and $e_j$. Inspired by the training of the KGE model, we obtain the normalized probability through the softmax function. Additionally, as there could be multiple valid tail entities for $(h,r,?)$, which should all obtain a probability close to $1$, hence we multiply the normalized probability by the number of tail entities and obtain:
\begin{equation}
\hat{p_r}\left(e_i,e_j\right)=\frac{exp{\left(f_r\left(e_i,e_j\right)\right)}\cdot\left|\left\{\left(e_i,r,e\right)\in\mathcal{T}\mid e\in\mathcal{V}\right\}\right|}{\sum_{e\in\mathcal{V}} exp{\left(f_r\left(e_i,e\right)\right)}}
\end{equation}
In order to make the neural adjacency matrix to more suitable for complex query answering tasks, we perform additional calibration on it. We learn an additional adaptation function $\phi_c$, which is parameterized by $\theta_1=\{\alpha,\beta\}$, and subsequently apply it to $\hat{p_r}\left(e_i,e_j\right)$, such that:
\begin{equation}
\phi_c\left(\hat{p_r}\left(e_i,e_j\right)\right)=\hat{p_r}\left(e_i,e_j\right)\left(1+\alpha_{i,r}\right)+\beta_{i,r}
\end{equation}
where $\alpha\in\mathbb{R}^{|\mathcal{V}|\times|\mathcal{R}|}$ and $\beta\in\mathbb{R}^{|\mathcal{V}|\times|\mathcal{R}|}$, with $\alpha_{i,r}\in\alpha$ and $\beta_{i,r}\in\beta$, are parameters of the adaptation function $\phi_c$. \\
Finally, we round the adapted calibration result so that it remains between $[0,1]$ and is consistent with real triplets in the KG:
\begin{equation}
p_r\left(e_i,e_j\right)=
\left\{
\begin{array}{ll}
1, &\left(r_i,r,e_j\right)\in\mathcal{T} \\
g(e_i, e_j), & otherwise
\end{array}
\right.
\end{equation}
where $g(e_i, e_j) = max\left\{min\left\{\phi_c\left(\hat{p_r}\left(e_i,e_j\right)\right),1-\delta \right\},\delta\right\}$, and we set $\delta=0.0001>0$ to avoid over-confidence on prediction.

For a given a KG, we use a pretrained KGE model to pre-compute its neural adjacency matrix $\mathbf{M}$, which is then stored for efficient query answering. As the computational complexity of pre-computing scales with the number of entities and relations in the KG, we predict what relations each entity may possess based on type information in the KG during computation. This serves to significantly reduce the pre-computing time required for the neural adjacency matrix. Specifically, if there is $\left(h,r\right)\in\mathcal{T}_{hr}$ in $\mathcal{G}_{hr}=\left(\mathcal{V},\mathcal{R},\mathcal{T}_{hr}\right)$, the KGE model pre-computes $\left(\mathbf{M}_r\right)_h$, which represents all the probability of the triplet $\left(h, r, ?\right)$ being true, otherwise skipping the computation and assigning $\left(\mathbf{M}_r\right)_h=\mathbf{0}$. Besides, considering that there are a large number of zero entries in $\mathbf{M}$, we optimize storage efficiency by adopting the sparse matrix storage technique, which greatly reduces storage space. With a type-based head entity-relation graph, we effectively avoid redundant computations, and reduce the overhead of pre-computing, including computing time, memory and storage space.

\subsection{Type-based Neural Link Prediction Adapter}
We propose a type-based neural link prediction adapter, which adjusts the neural link predictor based on type information by learning an adaptation function $\phi_a$. Let $\mathbf{T}=\left[p_1,p_2,\cdots,p_{\left|\mathcal{V}\right|}\right]\in\left[0,1\right]^{\left|V\right|}$ denote the output of neural link predictor, if there exists $\left(e_j,r\right)\in\mathcal{T}_{tr}$ in $\mathcal{G}_{tr}=\left(\mathcal{V},\mathcal{R},\mathcal{T}_{tr}\right)$, then $e_j$ is regarded as a relevant tail entity of $r$ based on types. The parameters $\theta_2=\{\gamma,\mu\}$ are used to adjust tail entities adaptively:
\begin{equation}
q_r\left(e_i,e_j\right)=
\left\{
\begin{array}{ll}
t(e_i, e_j), & if\left(e_j,r\right)\in\mathcal{T}_{tr} \\
p_j, & otherwise
\end{array}
\right.
\end{equation}
where $t(e_i, e_j)=max\left\{min\left\{p_j\left(1+\gamma_r\right)+\mu_r,1 \right\},0\right\}$, $\gamma\in\mathbb{R}^{|\mathcal{R}|}$ and $\mu\in\mathbb{R}^{|\mathcal{R}|}$.

\begin{figure*}[htb]
    \centering 
    \includegraphics[width=0.95 \linewidth]{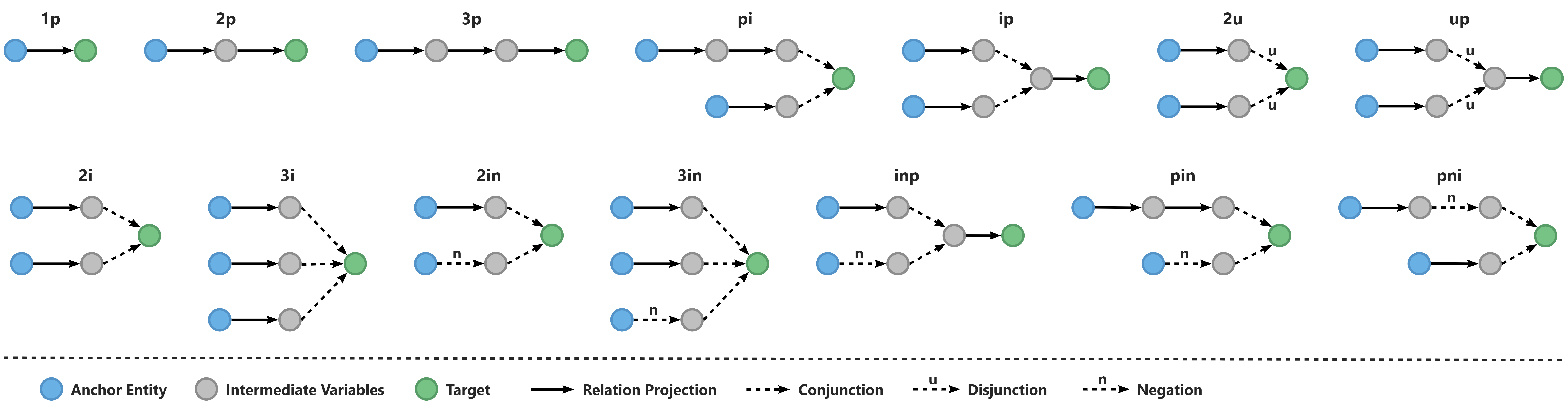}
    \caption{Query structures. There is a total number of 14 query types.}
    \label{figure3}
\end{figure*}

Intuitively, all valid head entities for $(?,r,t)$ should share a common class and possess similar or identical type information, likewise all valid tail entities. As described in Section \ref{type:graph}, we construct an entity-relation graph $\mathcal{G}_{tr}$ to incorporate type information into the learning of the adaptation function $\phi_a$. According to type information, TENLPA adaptively adjusts the neural link prediction results of specific entities to enhance the accuracy and reliability of predictions. Especially, TENLPA significantly improves on query tasks with negation, which is attributed to the auxiliary adjustment of type information. Let us consider the query \textit{“Who won the Turing Award?”}, which should clearly yield a set of names as its answer. Likewise, the negation of the query \textit{“Who did not win the Turing Award?”} should also produce a set of names. However, previous works only model the anti-relational projection through $\mathbf{1}-\mathbf{M}_r$ when executing the negation of queries. This results in high probabilities for many non-target types of entities in prediction results, such as cities, schools, films, and so forth. Fortunately, TENLPA effectively mitigates this issue by increasing the probability of target entities in prediction results through auxiliary supervision adjustment of type information, thereby distinguishing target type entities from non-target type entities and greatly improving the performance of the negation operation.

\subsection{Training}
Following previous works \citep{gnnqe}, our model is trained to minimize the binary cross entropy loss:
\begin{equation}
\begin{split}
\mathcal{L}=&-\frac{1}{\left|\mathcal{A}_Q\right|}\sum_{a\in\mathcal{A}_Q}\log p\left(a\mid Q\right) \\
&-\frac{1}{\left|\mathcal{V}\setminus\mathcal{A}_Q\right|}\sum_{a^\prime\in\mathcal{V}\setminus\mathcal{A}_Q}\log\left(1-p\left(a^\prime\mid Q\right)\right)    
\end{split}
\end{equation}
where $\mathcal{A}_Q$ is the set of answers to the complex query $Q$ and $p(a\mid Q)$ is the probability of entity $a$ in the final output result. Since TENLPA always outputs probabilities for all entities, we do not perform negative sampling and instead compute the loss with all negative answers.

\section{EXPERIMENTS}
In this section, we conduct experiments to demonstrate the effectiveness and efficiency of TENLPA on the task of answering complex queries. We discuss the experiment results and conduct further analysis on TENLPA. We systematically evaluate the performance of TENLPA on three standard benchmark datasets and investigate the impact of adaptive calibration and type-based adaptive adjustment mechanisms of our model by ablation experiments. We also analyse the space and time complexity to show the advantages of TENLPA.

\subsection{Experiment Setup}

\textbf{Datasets.} We conduct experiments on three knowledge graph datasets, including FB15k \citep{transe:fb15k}, FB15k-237 \citep{fb237} and NELL995 \citep{nell995}. We use the standard FOL queries generated in BetaE \citep{betae}, which consist of 9 types of EPFO queries (1p/2p/3p/2i/3i/pi/ip/2u/up) and 5 types of queries with negation (2in/3in/inp/pin/pni). In the query structure, “p”, “i” and “u” represent “projection”, “intersection” and “union”, respectively. We evaluate the model on all query types.  Detailed statistics of the datasets and the query types are shown in Table \ref{table1} and Figure \ref{figure3}. During training, we only use $\le 35\%$ of the training dataset (2i,3i,2in,3in) for the neural link prediction adapter.

\begin{table}[htb]
  \caption{Statistics on the different types of query structures.}
  \label{table1}
  \begin{center}
  \resizebox{\linewidth}{!}{
  \begin{tabular}{ccccc}
    Split&Query Type&FB15K&FB15K-237&NELL995 \\
    \hline \\
    \multirow{2}{*}{Training}& 1p,2p,3p,2i,3i & 273,710 & 149,689 & 107,982 \\
                             & Others & 27,371 & 14,968 & 10,798  \\
    \multirow{2}{*}{Validation} & 1p & 59,078 & 20,094 & 16,910  \\
                                & Others & 8,000 & 5,000 & 4,000   \\
    \multirow{2}{*}{Test}       & 1p & 66,990 & 22,804 & 17,021  \\
                                & Others & 8,000 & 5,000 & 4,000   \\ 
  \end{tabular}
  }
  \end{center}
\end{table}

\begin{table*}[htb]
  \caption{MRR results for answering queries on the testing sets. $\rm avg_p$ is the average on EPFO queries. $\rm avg_{ood}$ is the average on out-of-distribution (OOD) queries. $\rm avg_n$ is the average on queries with negation.}
  \label{table2}
  \begin{center}
  \resizebox{\linewidth}{!}{
  \begin{tabular}{cccccccccccccccccc}
    Model&$\rm avg_p$&$\rm avg_{ood}$&$\rm avg_n$&1p&2p&3p&2i&3i&pi&ip&2u&up&2in&3in&inp&pin&pni \\
    \hline \\
    \multicolumn{18}{c}{FB15K} \\                                              
    GQE+TEMP       & 46.6 & 39.6 & - & 74.9 & 31.4 & 26.0 & 59.3 & 69.4 & 47.3 & 35.9 & 47.8 & 27.4 & - & - & - & - & - \\
    Q2B+TEMP       & 44.0 & 34.8 & - & 74.8 & 25.6 & 22.3 & 61.7 & 72.6 & 43.7 & 29.0 & 44.1 & 22.5 & - & - & - & - & - \\
    BetaE+TEMP     & 44.6 & 37.4 & 12.5 & 70.3 & 28.9 & 25.8 & 58.2 & 68.4 & 45.8 & 32.2 & 44.3 & 27.3 & 15.2 & 15.6 & 11.5 & 6.8 & 13.4 \\
    CQD-CO   & 46.9 & 35.3   & -    & 89.2 & 25.3 & 13.4 & 74.4 & 78.3 & 44.1 & 33.2 & 41.8 & 21.9 & -    & -    & -    & -    & -    \\
    CQD-Beam & 58.2 & 49.8   & -    & 89.2 & 54.3 & 28.6 & 74.4 & 78.3 & 58.2 & 67.7 & 42.4 & 30.9 & -    & -    & -    & -    & -    \\
    GNN-QE    & 72.8 & 68.9   & 38.6 & 88.5 & 69.3 & 58.7 & 79.7 & 83.5 & 69.9 & 70.4 & 74.1 & 61.0 & 44.7 & 41.7 & 42.0 & 30.1 & \textbf{34.3}  \\
    QTO       & 74.0 & 71.8   & 49.2 & 89.5 & 67.4 & 58.8 & \textbf{80.3} & 83.6 & 75.2 & 74.0 & \textbf{76.7} & 61.3 & 61.1 & 61.2 & 47.6 & 48.9 & 27.5  \\
    TENLPA    & \textbf{74.2} & \textbf{71.9} & \textbf{50.9} & \textbf{89.5} & \textbf{67.6} & \textbf{59.0} & 80.2 & \textbf{84.3} & \textbf{75.2} & \textbf{74.3} & 76.3          & \textbf{61.8} & \textbf{62.2} & \textbf{64.2} & \textbf{48.3} & \textbf{50.3} & 29.6  \\
    \multicolumn{18}{c}{FB15K-237} \\                                          
    GQE+TEMP       & 23.7 & 16.6 & - & 42.9 & 12.3 & 10.1 & 34.4 & 47.6 & 26.0 & 15.1 & 15.1 & 10.1 & - & - & - & - & - \\
    Q2B+TEMP       & 22.1 & 13.9 & - & 40.9 & 11.0 & 9.2 & 33.7 & 48.2 & 21.4 & 12.3 & 12.9 & 9.1 & - & - & - & - & - \\
    BetaE+TEMP     & 22.5 & 15.3 & 5.3 & 39.9 & 11.8 & 10.5 & 32.6 & 46.7 & 24.9 & 13.6 & 12.5 & 10.2 & 4.3 & 8.0 & 7.6 & 3.5 & 2.9 \\
    CQD-CO   & 21.8 & 15.6   & -    & 46.7 & 9.5  & 6.3  & 31.2 & 40.6 & 23.6 & 16.0 & 14.5 & 8.2  & -    & -    & -    & -    & -    \\
    CQD-Beam & 22.3 & 15.7   & -    & 46.7 & 11.6 & 8.0  & 31.2 & 40.6 & 21.2 & 18.7 & 14.6 & 8.4  & -    & -    & -    & -    & -    \\
    GNN-QE    & 26.8 & 19.9   & 10.2 & 42.8 & 14.7 & 11.8 & 38.3 & 54.1 & 31.1 & 18.9 & 16.2 & 13.4 & 10.0 & 16.8 & 9.3  & 7.2  & \textbf{7.8}  \\
    QTO       & 33.5 & 27.6   & 15.5 & 49.0 & 21.4 & 21.2 & 43.1 & 56.8 & 38.1 & 28.0 & 22.7 & \textbf{21.4} & 16.8 & 26.7 & 15.1 & 13.6 & 5.4  \\
    TENLPA    & \textbf{33.8} & \textbf{27.8} & \textbf{16.4} & \textbf{49.1} & \textbf{21.6} & \textbf{21.4} & \textbf{43.3} & \textbf{57.6} & \textbf{38.5} & \textbf{28.5} & \textbf{23.0} & 21.3          & \textbf{17.7} & \textbf{28.3} & \textbf{16.0} & \textbf{14.3} & 5.9  \\
    \multicolumn{18}{c}{NELL995}  \\
    GQE+TEMP       & 28.0 & 18.3 & - & 57.7 & 17.2 & 14.1 & 40.6 & 49.9 & 27.0 & 18.5 & 15.9 & 11.6 & - & - & - & - & - \\
    Q2B+TEMP       & 26.4 & 15.2 & - & 56.5 & 15.0 & 12.9 & 40.8 & 52.0 & 21.1 & 16.0 & 14.2 & 9.4 & - & - & - & - & - \\
    BetaE+TEMP     & 25.5 & 15.5 & 5.9 & 54.1 & 14.2 & 12.4 & 38.1 & 48.9 & 23.9 & 16.0 & 12.8 & 9.2 & 5.1 & 7.5 & 10.5 & 3.1 & 5.9 \\
    CQD-CO   & 28.8 & 20.7   & -    & 60.4 & 17.8 & 12.7 & 39.3 & 46.6 & 30.1 & 22.0 & 17.3 & 13.2 & -    & -    & -    & -    & -    \\
    CQD-Beam & 28.6 & 19.8   & -    & 60.4 & 20.6 & 11.6 & 39.3 & 46.6 & 25.4 & 23.9 & 17.5 & 12.2 & -    & -    & -    & -    & -    \\
    GNN-QE    & 28.9 & 19.6   & 9.7  & 53.3 & 18.9 & 14.9 & 42.4 & 52.5 & 30.8 & 18.9 & 15.9 & 12.6 & 9.9  & 14.6 & 11.4 & 6.3  & 6.3  \\
    QTO       & 32.9 & 24.0   & 12.9 & 60.7 & 24.1 & 21.6 & 42.5 & 50.6 & 31.3 & 26.5 & 20.4 & 17.9 & 13.8 & 17.9 & 16.9 & 9.9  & 5.9  \\
    TENLPA    & \textbf{33.8} & \textbf{24.8} & \textbf{13.1} & \textbf{60.8} & \textbf{24.1} & \textbf{21.8} & \textbf{44.2} & \textbf{54.4} & \textbf{32.9} & \textbf{27.6} & \textbf{20.4} & \textbf{18.2} & \textbf{14.1} & \textbf{18.1} & \textbf{17.1} & \textbf{10.0}          & \textbf{6.0}  \\
  \end{tabular}
  }
  \end{center}
\end{table*} 

\textbf{Baselines.} We compare TENLPA with state-of-the-art methods on complex query answering. Specifically, we choose GQE \citep{gqe}, Query2Box \citep{q2b}, and BetaE \citep{betae} with TEMP \citep{temp} as strong baselines for embedding methods. We also compare with methods that utilize Fuzzy Logic and GNNs for enhancing the properties of the embedding space, such as GNN-QE \citep{gnnqe}. Finally, we evaluate our method against CQD-CO \citep{cqd}, CQD-Beam \citep{cqd} and the state-of-the-art QTO \citep{qto} that leverage neural link predictors.

\textbf{Evaluation Protocol.} Following the evaluation protocol in BetaE \citep{betae}, we separate the answers to each query into two sets: easy answers and hard answers. For test (validation) queries, easy answers refer to entities that can be reached by edges in training/validation graph, while hard answers are those that can only be inferred by predicting missing edges in the valid/test graph. The performance is calculated by standard evaluation metrics, namely mean reciprocal rank (MRR) and Hits at K (Hits@K) on hard answers.

\textbf{Implementation Details.} 
We employ ComplEx-N3-RP \citep{complex,n3,rp}, the current state-of-the-art KGE methods as the KGE model, followed by computation and calibration of the neural adjacency $\mathbf{M}$. Referring to the hyperparameter settings of QTO \citep{qto}, we train the adaptive functions in our model using Adagrad as an optimizer, with grid search to find the best hyperparameter settings on the validation set. The learning rate is set to be $10^{-5}$, and will decrease during the training process. 

\subsection{Main Results}
Table \ref{table2} shows the MRR results of different models for answering complex logical queries. GQE, Q2B, CQD-CO, and CQD-Beam do not support queries with negation, so the corresponding entries are empty. Except CQD and QTO, previous baselines are trained on 1p/2p/3p/2i/3i queries, so we regard the other 4 types of EPFO queries as OOD queries and report the average results on these queries in $\rm avg_{ood}$. We observe that TENLPA outperforms baseline methods significantly across all query types and datasets. TENLPA yields a relative gain of 1.3\%, 1.4\% and 3.6\% on $\rm avg_p$, $\rm avg_{ood}$, $\rm avg_n$ over previous state-of-the-art QTO, which implies that our method demonstrates better reasoning skills and superior adaptability when tackling complex query answering tasks. We attribute this gain to the integration of type information and the benefit of adaptive tuning.

\begin{figure}[htb]
    \centering 
    \includegraphics[width=0.95 \linewidth]{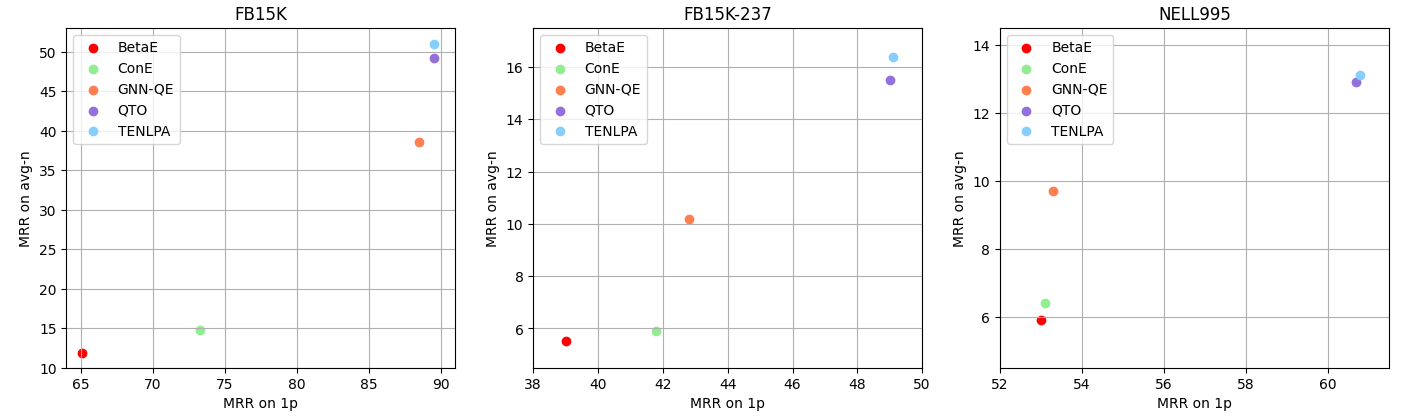}
    \caption{MRR results on queries with negation w.r.t. MRR results on knowledge graph completion (1p queries).}
    \label{figure2}
\end{figure}

\begin{table*}[htb]
  \caption{MRR results for ablation study on the testing sets. $\rm avg_p$ is the average on EPFO queries. $\rm avg_{ood}$ is the average on out-of-distribution (OOD) queries. $\rm avg_n$ is the average on queries with negation.}
  \label{table3}
  \resizebox{\linewidth}{!}{
  \begin{tabular}{cccccccccccccccccc}
    Model&$\rm avg_p$&$\rm avg_{ood}$&$\rm avg_n$&1p&2p&3p&2i&3i&pi&ip&2u&up&2in&3in&inp&pin&pni \\
    \hline \\
    \multicolumn{18}{c}{FB15K} \\                                              
    QTO      & 74.0          & 71.8          & 49.2          & 89.5          & 67.4          & 58.8          & 80.3          & 83.6          & 75.2          & 74.0          & \textbf{76.7} & 61.3          & 61.1          & 61.2          & 47.6          & 48.9          & 27.5          \\
    TENLPA$^{\rm-t}$ & 74.2          & 71.9          & 49.0          & 89.5          & 67.6          & 59.0          & 80.2          & 84.3          & 75.2          & 74.3          & 76.3          & 61.8          & 60.5          & 60.5          & 48.3          & 48.4          & 27.1          \\
    TENLPA$^{\rm-c}$ & 74.0          & 71.7          & 50.9          & 89.5          & 67.6          & 58.9          & 79.8          & 82.9          & 74.9          & 73.9          & 76.3          & 61.8          & 62.2          & 64.2          & 48.3          & 50.3          & 29.6          \\
    TENLPA   & \textbf{74.2} & \textbf{71.9} & \textbf{50.9} & \textbf{89.5} & \textbf{67.6} & \textbf{59.0} & \textbf{80.2} & \textbf{84.3} & \textbf{75.2} & \textbf{74.3} & 76.3          & \textbf{61.8} & \textbf{62.2} & \textbf{64.2} & \textbf{48.3} & \textbf{50.3} & \textbf{29.6} \\
    \multicolumn{18}{c}{FB15K-237} \\                                          
    QTO      & 33.5          & 27.6          & 15.5          & 49.0          & 21.4          & 21.2          & 43.1          & 56.8          & 38.1          & 28.0          & 22.7          & \textbf{21.4} & 16.8          & 26.7          & 15.1          & 13.6          & 5.4           \\
    TENLPA$^{\rm-t}$ & 33.8          & 27.8          & 15.6          & 49.1          & 21.6          & 21.4          & 43.3          & 57.6          & 38.5          & 28.5          & 23.0          & 21.3          & 16.9          & 26.7          & 15.1          & 13.6          & 5.5           \\
    TENLPA$^{\rm-c}$ & 33.7          & 27.8          & 16.4          & 49.1          & 21.6          & 21.3          & 43.0          & 57.0          & 38.3          & 28.4          & 23.0          & 21.3          & 17.7          & 28.3          & 16.0          & 14.3          & 5.9           \\
    TENLPA   & \textbf{33.8} & \textbf{27.8} & \textbf{16.4} & \textbf{49.1} & \textbf{21.6} & \textbf{21.4} & \textbf{43.3} & \textbf{57.6} & \textbf{38.5} & \textbf{28.5} & \textbf{23.0} & 21.3          & \textbf{17.7} & \textbf{28.3} & \textbf{16.0} & \textbf{14.3} & \textbf{5.9}  \\
    \multicolumn{18}{c}{NELL995}  \\
    QTO      & 32.9          & 24.0          & 12.9          & 60.7          & 24.1          & 21.6          & 42.5          & 50.6          & 31.3          & 26.5          & 20.4          & 17.9          & 13.8          & 17.9          & 16.9          & 9.9           & 5.9           \\
    TENLPA$^{\rm-t}$ & 33.8          & 24.8          & 12.8          & 60.8          & 24.0          & 21.6          & 44.2          & 54.4          & 32.9          & 27.6          & 20.4          & 18.2          & 13.8          & 17.6          & 17.1          & 9.7           & 6.0           \\
    TENLPA$^{\rm-c}$ & 32.9          & 24.2          & 13.0          & 60.8          & 24.1          & 21.8          & 42.5          & 50.3          & 31.4          & 26.8          & 20.4          & 18.2          & 14.1          & 18.1          & 16.9          & 10.0          & 6.0           \\
    TENLPA   & \textbf{33.8} & \textbf{24.8} & \textbf{13.1} & \textbf{60.8} & \textbf{24.1} & \textbf{21.8} & \textbf{44.2} & \textbf{54.4} & \textbf{32.9} & \textbf{27.6} & \textbf{20.4} & \textbf{18.2} & \textbf{14.1} & \textbf{18.1} & \textbf{17.1} & \textbf{10.0} & \textbf{6.0} \\
\end{tabular}
}
\end{table*}

Compared with pure embedding methods like GQE, Q2B and BetaE, TENLPA achieves an impressive improvement on all types of queries. This indicates that fuzzy sets have advantages in modeling logical operations. Despite GNN-QE also use fuzzy sets to model intermediate variables with many possible assignments, our model have better capabilities due to a more powerful one-hop answering KGE model. Moreover, we can see that CQD generalizes worse than our model, indicating our optimization method better leverages the pretrained KGE model.

Intuitively, the performance of complex query models should benefit from better KG completion performance, i.e., 1p queries. Here we disentangle the contribution of KG completion and complex query framework in answering queries with negation. Figure \ref{figure2} plots the performance on queries with negation w.r.t. the performance on 1p queries on all datasets. The point of TENLPA lies on the top left to baseline methods, illustrating it generalizes better from KG completion to complex queries. TENLPA shows a better generalization from KG completion to queries with negation, which suggests its adaptive mechanism is useful and critical.

\subsection{Ablation Study}

To evaluate the significance of TENLPA’s two main modules, adaptive calibration mechanism and type-based adaptive adjustment mechanism, we conduct two ablation studies by removing any one module from the full model on all three datasets. Since TENLPA is an extension of QTO, we can consider QTO as the basic model with both two modules removed. Table \ref{table2} reports the MRR results for answering complex queries, where $\rm TENLPA^{-t}$ represents the results of TENLPA without type-based adaptive adjustment mechanism, and $\rm TENLPA^{-c}$ represents the results of TENLPA without adaptive calibration mechanism. 

It can be seen that adaptive calibration mechanism and type-based adaptive adjustment mechanism both contribute to the accurate result of our complete model. We make two key observations: (a) $\rm TENLPA^{-t}$ mainly improves the results on $\rm avg_p$ with a relative gain of 1.3\%, suggesting adaptive calibration mechanism plays an important role in EPFO queries through re-calibrating adjacency matrix; (b) $\rm TENLPA^{-c}$ achieves the significant improvement on $\rm avg_n$ with a comparative increase of 3.6\%, indicating type-based adaptive adjustment mechanism successfully leverages type information to better answer queries with negation.Through comparison, our experiment results intuitively demonstrate that the two modules have their own emphasis on improving the performance of the model, and both are reliable and effective.

\subsection{Discussion}
\textbf{Space Complexity.} We first consider the storage usage of TENLPA. The neural adjacency matrix $\mathbf{M}$ contains $\left|\mathcal{R}\right|\cdot\left|\mathcal{V}\right|^2$ entries. During computation, we incorporate type information to predict the relation that each entity may possess. If $h\in\mathcal{V}$ does not possess $r\in\mathcal{R}$, all the entries correlated to it are assigned a value of $0$. Meanwhile, because of the sparsity of the KG, most entries in $\mathbf{M}$ have relatively small values that can be filtered through an appropriate threshold $\epsilon>0$, while maintaining precision. Compared to QTO, TENLPA can further optimize the storage usage of $\mathbf{M}$, with a reduced storage usage only 60-75\% of QTO.

\textbf{Time Complexity.} Moreover, we evaluate the time complexity of TENLPA. With the incorporation of type information, we can avoid redundant computations and significantly reduce the pre-computing time. In comparison to QTO, TENLPA demonstrates a faster processing speed, with the reduced pre-computing time of approximately 20\%.

\section{CONCLUSION}
Answering complex logical queries on KGs is a fundamental and significant task that has always faced challenges stemming from factors such as missing information and noise. In this paper, we propose a novel model, namely Type-based Neural Link Prediction Adapter (TENLPA), for answering complex logical queries on KGs. This model presents an adaptive mechanism with type information auxiliary over neural link predictors, which does not require data-intensive training. TENLPA supports all the FOL operations and can be trained with only a subset of query types. Experiments demonstrate that TENLPA outperforms previous methods, especially on queries with negation. Future work may expand the application fields of TENLPA, such as answering multi-hop questions in natural language and complex logical query in specialized domains.

\setlength{\itemindent}{-\leftmargin}
\makeatletter\renewcommand{\@biblabel}[1]{}\makeatother
\bibliographystyle{abbrvnat}
\bibliography{sample-base}

\end{document}